\documentclass[twocolumn,showpacs,prl]{revtex4-1}
\usepackage{amsmath}
\usepackage{amsfonts}
\usepackage{graphicx}
\usepackage{color}
\usepackage{comment}

\usepackage{algorithm}
\usepackage{algorithmic}
\usepackage{hyperref}
\usepackage{natbib}

\begin{document}

\title{Counterfactual Control for Free from Generative Models}
\author{Nicholas Guttenberg}
\affiliation{Araya, Tokyo}
\affiliation{Earth-Life Science Institute, Tokyo, Japan}
\author{Yen Yu}
\author{Ryota Kanai}
\affiliation{Araya, Tokyo}

\begin{abstract}
We introduce a method by which a generative model learning the joint distribution between actions and future states can be used to automatically infer a control scheme for any desired reward function, which may be altered on the fly without retraining the model. In this method, the problem of action selection is reduced to one of gradient descent on the latent space of the generative model, with the model itself providing the means of evaluating outcomes and finding the gradient, much like how the reward network in Deep Q-Networks (DQN) provides gradient information for the action generator. Unlike DQN or Actor-Critic, which are conditional models for a specific reward, using a generative model of the full joint distribution permits the reward to be changed on the fly. In addition, the generated futures can be inspected to gain insight in to what the network 'thinks' will happen, and to what went wrong when the outcomes deviate from prediction.
\end{abstract}

\maketitle

\section{Introduction}
In standard reinforcement learning, control policies are usually optimized to maximize expected future rewards\cite{sutton1998reinforcement}, and the learning of the structure in the environment is grounded to pre-defined goals or rewards. However, in a complex environment, rewards are often sparse and delayed, and even what constitutes as a reward remains unclear until a specific task is given. Furthermore, for the embedded agent, the environment's reaction to the agent's actions is a black box process, and the gradient information for control policy needs to be obtained indirectly from feedback on success or failure. Previous studies have addressed this issue by using a known model for the environment\cite{sutton1991dyna, doya2002multiple}, through sampling\cite{williams1992simple,moriarty1999evolutionary}, or by methods which learn to predict the reward which will follow from a given action or policy\cite{mnih2015human, mnih2016asynchronous}. Alternately, given examples of correct behavior from an expert, a model can be made which simply tries to predict what the expert would do in that situation\cite{abbeel2004apprenticeship}. 

Most model-learning methods focus on predictive models\cite{lenz2015deepmpc,oh2015action,fragkiadaki2015learning,dosovitskiy2016learning}: given that some state occurred, and some action was taken, predict what reward resulted ($p(reward|action,state)$). Action-conditional prediction models \cite{oh2015action,fragkiadaki2015learning} learning to predict future images conditioned on actions amounts to learning a model of the dynamics of the agent-environment interaction, an essential component of model-based approaches to reinforcement learning.

This paper proposes an alternative approach in which control policies are not directly learned through rewards or for specific goals, but are derived on the fly from generative models of the joint distribution between future state and action that an agent has learned through interactions with the environment. Once an agent has acquired a generative model of possible future action-state trajectories, it can generate counterfactual futures to evaluate accessible future outcomes to plan action sequences in any task. That is to say, the generative model of future state and action $p(future, action | state)$ can be used for creating a control policy for any desired statistics that the agent aims to achieve in the future. This property is desirable when we are interested in implementing intrinsic motivation \cite{schmidhuber2010formal} to minimize the entropy of hidden states in the model \cite{houthooft2016vime}, which has been conceived as a possible way to formulate saliency in visual search \cite{friston2012perceptions} and to encourage sampling from  space where information gain is expected to be high.  

In contrast to the action-conditional prediction models, we expect that our approach lead to a more general result. This is because one may obtain all of the component conditional distributions given the joint distribution, but not vice versa. In other words, the joint distribution $p(x,y)$ can be used without modification to generate conditional probabilities over any combination of the variables, which allows evaluation of both $p(x|y)$ and $p(y|x)$. 
The generative model can imagine counterfactual sequences of events that are internally consistent (samples from $p(action,future|state)$), evaluate any desired reward function on that imagined future, and then choose to follow the action sequence with the best anticipated reward. Because the model generates rich future information, the reward function is separated from learning the model (at least, to the extent that the data set covers that portion of the state space), meaning that the reward function can be changed on the fly without additional training. As such, the model learns the affordances of its environment and can then exploit those as necessary (via sampling or search), rather than learning a method for solving a specific task. Inverse reinforcement learning results demonstrate that the conditional statistics of an expert's behaviors are sufficient to reconstruct successful policies \cite{ho2016generative}, suggesting that perhaps a generative model connecting behavior and outcome rather than behavior and state might itself contain sufficient information to select good actions even in the absence of expert examples. 

Furthermore, it is easier to build reward functions which include uncertainty estimates as part of the reward if one is starting from the joint distribution, rather than sampling individual runs. These analyses suggest that unlike the action-conditional prediction models, the joint generative models assure that action sequences are consistent with state sequence. Additionally, action and state sequences are forced to be more plausible as higher order correlations are directly captured in the joint distribution. 

In this paper, we explore the idea that learning a generative model of action and state, not directly trained to solve a particular task, can be used for inducing control policies for ad hoc goals. To demonstrate this concept, we aimed to show that 1) generative controllers are capable of controlling tasks and 2) that generative models can produce control policies for novel tasks, and thus showing the property of generalization.

\section{Architectural Considerations}
We motivate our choice of model architecture by calling upon the notion of counterfactual reasoning, with an attempt to disentangle recognition of action opportunities from implementation of a specific behavioral programme.

Action opportunities can be taken as latent attributes of perceptual cue if only the nature of embedded actor is specified. The term "opportunity" here appeals to the notion that the perceptual cue portends accessibility of plausible sensory consequence under action, with the action being purely fictive. This implies a model latent space that jointly represents action-outcome cohesion. Namely, a model that learns to predict the joint density $p(\{a, s\}|s_1)$ of fictive action-state pairs $\{a, s\}=\{a_2, s_2, a_3, \dots, a_n, s_n\}$ given initial state $s_1$. The training set may come from an unbiased sampling process or from an expert, but the learning algorithm will not be tailored to maximize any score function nonetheless. A model of the sort thus entails some latent distribution $z\sim q(z)$ encoding actions and sensory consequences, that one may perform inference (or search) on the latent space to satisfy some external enabler (e.g., a reward or goal function, $g$) using gradient descent, $\nabla_z\langle g(s_n)\rangle_q$. 

A natural choice of model architecture that encodes joint probability is one of generative models.There are a number of techniques to learn generative models. We note several major categories, including variational auto-encoders \cite{kingma2013vae}, generative adversarial networks \cite{goodfellow2014gan} and, in particular, a hybrid approach \cite{guttenberg2016recurrent} of our own inspired by recent results that Monte Carlo sampling over the auto-encoder latent space improves the fidelity of the distribution \cite{creswell2016improving}, and by work on the correspondence between recurrent neural networks and conditional random fields \cite{zheng2015conditional}. Our 
hybrid approach replaces the Monte Carlo sampling procedure on a standard variational auto-encoder with instead training a recurrent auto-encoder end-to-end. The recurrent connections simulate the effects of Monte Carlo sampling, but maintain the ability to propagate gradients across the sampling procedure. As a result, the network learns a fixed-point structure corresponding to the data distribution. 

The question then is, how to practically extract this information? Sampling future states until finding a good one is expensive, but depending on the generative model, a more sophisticated form of search may be possible. Many methods for learning generative models work by specifying some latent space with a known distribution, and then learning a transform from that latent space into the data space. This means that different futures are parameterized by coordinates in a potentially low-dimensional, structured space, connected to the data space by a known, differentiable function. As such, gradient descent over the latent space may be an efficient replacement for sampling or search processes. Several methods have been proposed to influence the structure of the latent space and its correspondence to known factors about the data \cite{chen2016infogan,mathieu2016dr}, which means it may even be possible to learn a generative model in such a way that certain desired properties of the solution can be directly specified. For example, in controlling the motion of a robot arm, one could tie some of the latent variables to the future position of the robot arm, and thereby be able to directly specify that position by imposing a prior on the latent space, without needing any search or gradient descent at all. This marks the departure from standard reinforcement learning in which action-outcome contingency is framed by reward specification.

\section{Experiments}

We apply the idea of deriving control policies from a generative model in the context of the cartpole balancing task (AI Gym environments `Cartpole-v0' and `Cartpole-v1') \cite{brockman2016gym}. This task involves balancing a pole on a frictionless cart, which can move left and right. There are four continuous sensor inputs (the horizontal position and velocity of the cart, as well as the angle and angular velocity of the pole) and one discrete action output, which accelerates the cart with a fixed force to either the left or the right. A PID controller provided a linearization of the pole's dynamics around the unstable fixed point can directly solve the balancing task \cite{minorsky1922pid}, but other methods require some amount of training time to learn a model or action policy enabling the pole to be balanced. We note that this is not a representative use-case for deep neural controllers, but rather we focus on this task due to its computational simplicity allowing us to survey a greater range of parameters and architectures in order to better understand and quantify the characteristics of generative models as controllers.

\subsection{Model Architecture}

\begin{figure}[t]
 \includegraphics[width=\columnwidth]{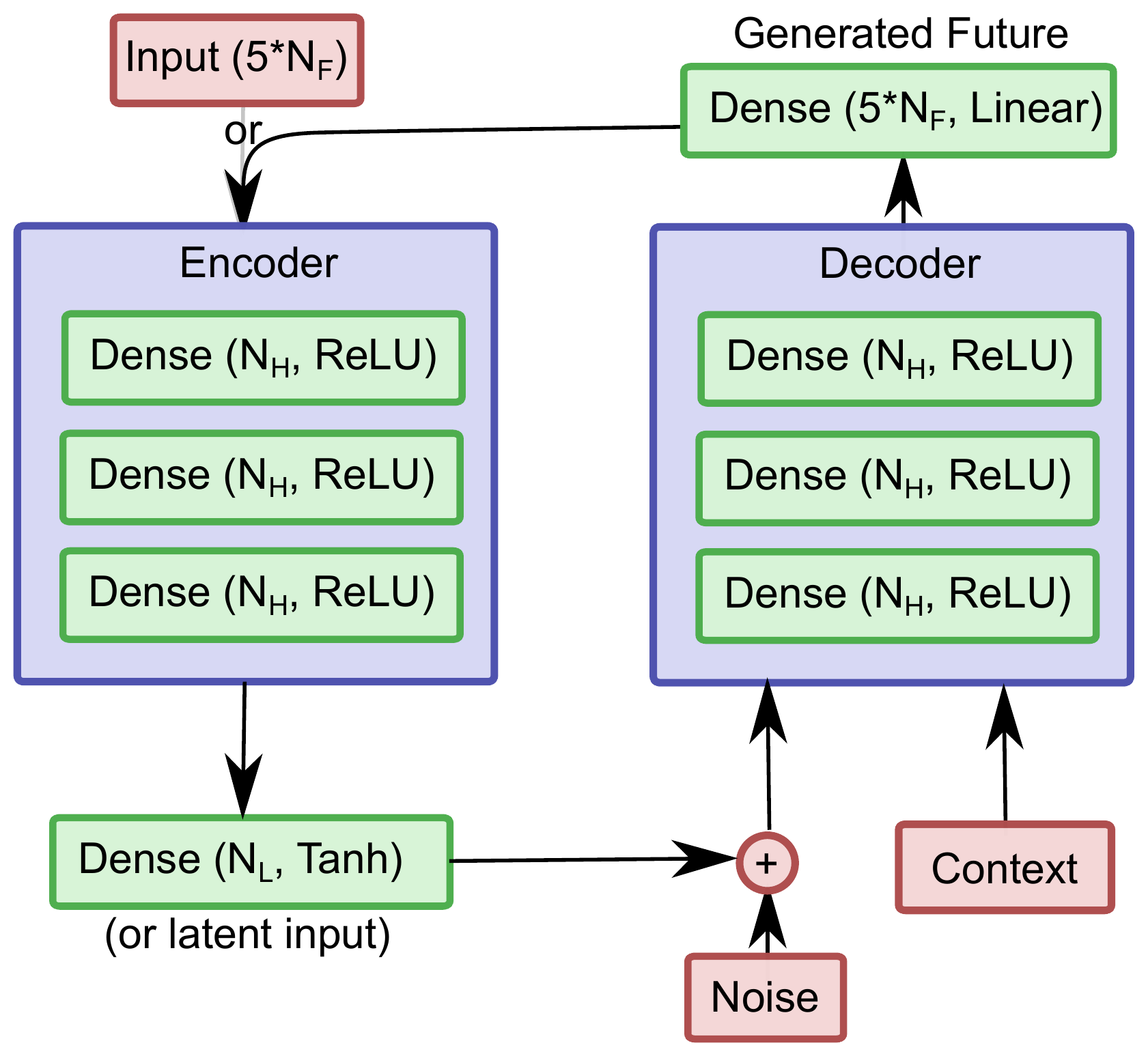}
 \caption{\label{Architecture}Recurrent auto-encoder architecture for generating future (action, state) sequences.}
\end{figure}

In our case, we will use a recurrent auto-encoder (e.g., \url{http://www.araya.org/archives/1306}) to learn a generative model of the next $N_F$ states and actions conditioned on the current state. We also attempted a similar setup as a GAN \cite{goodfellow2014gan}, but ultimately we encountered stability problems with the GAN that prevented the model from eventually converging to a successful action policy. We use a network with three hidden layers of size $N_H$ in the encoder, a latent space of size $N_L$ to which Gaussian noise with standard deviation $\sigma$ is added, and three hidden layers of size $N_H$ to decode back to the future (action, state) sequence. Conditioning on the current state is performed at the input to the decoder (that is, the same point in the latent space can be decoded differently depending on the current context). A schematic of this architecture is shown in Fig.~\ref{Architecture}.

During the task, we start with a random latent space vector and perform $n$ gradient descent steps to optimize the target reward function. We return the future (action, state) sequence, as well as the new latent space vector. This latent space vector is retained and used as the starting point for additional gradient descent steps whenever the policy is updated (which we do every other frame). We find this saves us about a factor of 10 in computation time compared to needing to gradient descend from scratch every frame. In addition to the reward function, we add an $L_2$ regularization to the gradient descent encouraging the latent vector to have a small magnitude --- this helps constrain the gradient descent to more `realistic' solutions. The full gradient descent step for a reward function $Q$ is:

$$\Delta Z = \alpha \frac{\partial_Z Q}{||\partial_Z Q||_1} - \beta Z$$

with $\alpha=0.05$ and $\beta = 0.001$. We perform 100 such steps at each update.

Sensor values as returned by the cartpole task have very different scales between position and velocity sensors, and so we multiply both the $x$ and $\theta$ sensors by a factor of 10 in their encoding order to make these more comparable. The actions are encoded to the network as either $-1$ or $+1$. The network's output, however, is not constrained to these values. In order to convert the network output into a can be anywhere in between these extreme values, we interpret the output $a$ as a probability $p_> = (a+1)/2$ of going to the right, and sample actions from that distribution.

\begin{table}
\begin{tabular}{|c|c|l|}
\hline
Parameter & Value & Description \\
\hline
$N_F$ & 16 & \# of future steps to generate \\
$N_L$ & 2 & Latent space dimension \\
$N_H$ & 256 & \# of hidden units \\
$\sigma$ & 0.2 & Noise added to latents \\
$N_r$ & 7 & \# of recurrent steps \\
\hline
$r$ & $10^{-4}$ & Learning rate \\
$N_t$ & 400 & \# of training steps per cycle \\
$N_e$ & 5 & \# of episodes per cycle \\
\hline
$\gamma_x$ & 1.2 & Reward weight of x position \\
$\gamma_v$ & 1.8 & Reward weight of x velocity \\
$\gamma_\theta$ & 3.0 & Reward weight of pole angle \\
$\gamma_\omega$ & 0.8 & Reward weight of $\omega$ \\
\hline
$N_g$ & 100 & \# of gradient descent steps \\
$\alpha$ & 0.05 & Gradient descent step size \\
$\beta$ & 0.001 & Latent space L2 norm \\
\hline
\end{tabular}
\caption{\label{Parameters}Hyperparameters for the cartpole generative controller.}
\end{table}

\subsection{Training directly}

First, we examine the behavior of the generative controller when being trained on data from its own attempts to solve a fixed task (balancing the pole). We accumulate data 5 episodes at a time, then perform 400 training steps before generating the next 5 episodes. Each training step consists of a batch of 1000 example sequences chosen randomly from a complete memory of all previous episodes. First the episode number is sampled uniformly, then the starting position within the episode is sampled uniformly, so this is not biased towards longer episodes. The network is trained using Adam \cite{kingma2014adam} with a learning rate of $10^{-4}$ and $\beta_1 = 0.5$. L1 regularization on the weights is added to the loss, with a scale of $5\times10^{-4}$. 

The reward function for Cartpole-v1 is simply the number of steps that the pole remains within a certain angle of the vertical, and does not move more than a certain amount horizontally from its initial position. For our generative controller, we need to express this reward (or something related to it) in terms of a differentiable function of the future projected state. Since this is evaluated at a specific point in time, we could end up with problems such as `the pole is in the right position, but moving very fast' if we were just to use the angle from the vertical. So we construct a reward function which encourages the pole to be near its starting position, near the vertical, and also not moving or rotating:

$$Q = \gamma_x |x-x_0| + \gamma_v v^2 + \gamma_\theta (\theta-\theta_0)^2 + \gamma_\omega \omega^2$$

\begin{figure}
 \includegraphics[width=\columnwidth]{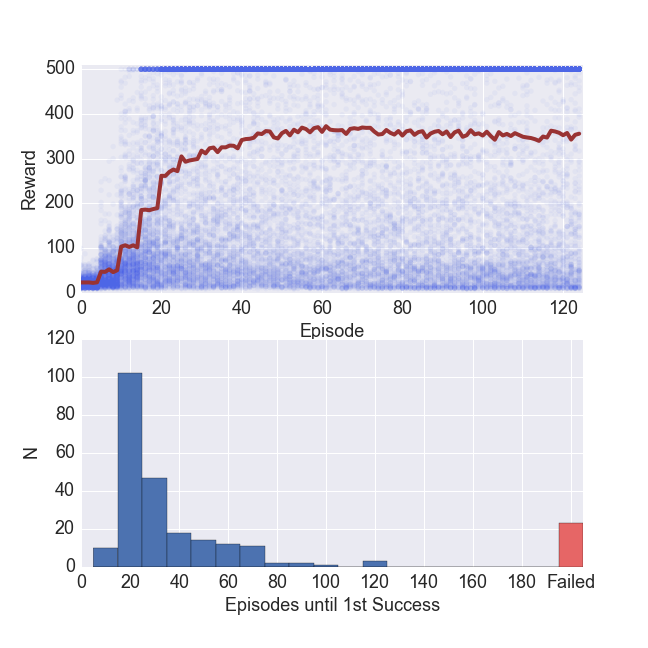}
 \caption{\label{Training}Top: Average training curve over 245 runs (line) and distribution (points). Maximum possible reward is 500. Bottom: Distribution of when the model first achieves a reward of 500. The average (over models which succeed during the 125 episode interval) is 32 episodes. 10\% of runs fail to solve the task within 125 episodes.}
\end{figure}

We were able to find a set of hyperparameters for which the generative controller can rapidly learn to balance the pole. These values are show in table \ref{Parameters}. For these values, in the best cases the model can learn to balance the pole for 500 steps after an average of 32 episodes of training. Furthermore, if trained against a longer balance time of 3000 steps, we found that the model trained to balance the pole in place was able to handle alternate reward functions such as a case in which $x_0$ varied sinusoidally with time. A few example training curves, along with the distribution of times until the model achieves its first perfect score are shown in Fig.~\ref{Training}. 

In terms of hyperparameters, although there are quite a few, in practice we found that one hyperparameter in particular was essential for obtaining a working controller - the size of the latent space $N_L$. If $N_L$ is too large, the model never became able to balance the pole significantly.  We suspect this is because when $N_L$ becomes large, there are more directions in the latent space that the gradient descent procedure can exploit to force a rewarding (but unrealistic) future prediction. 

\begin{figure}
 \includegraphics[width=\columnwidth]{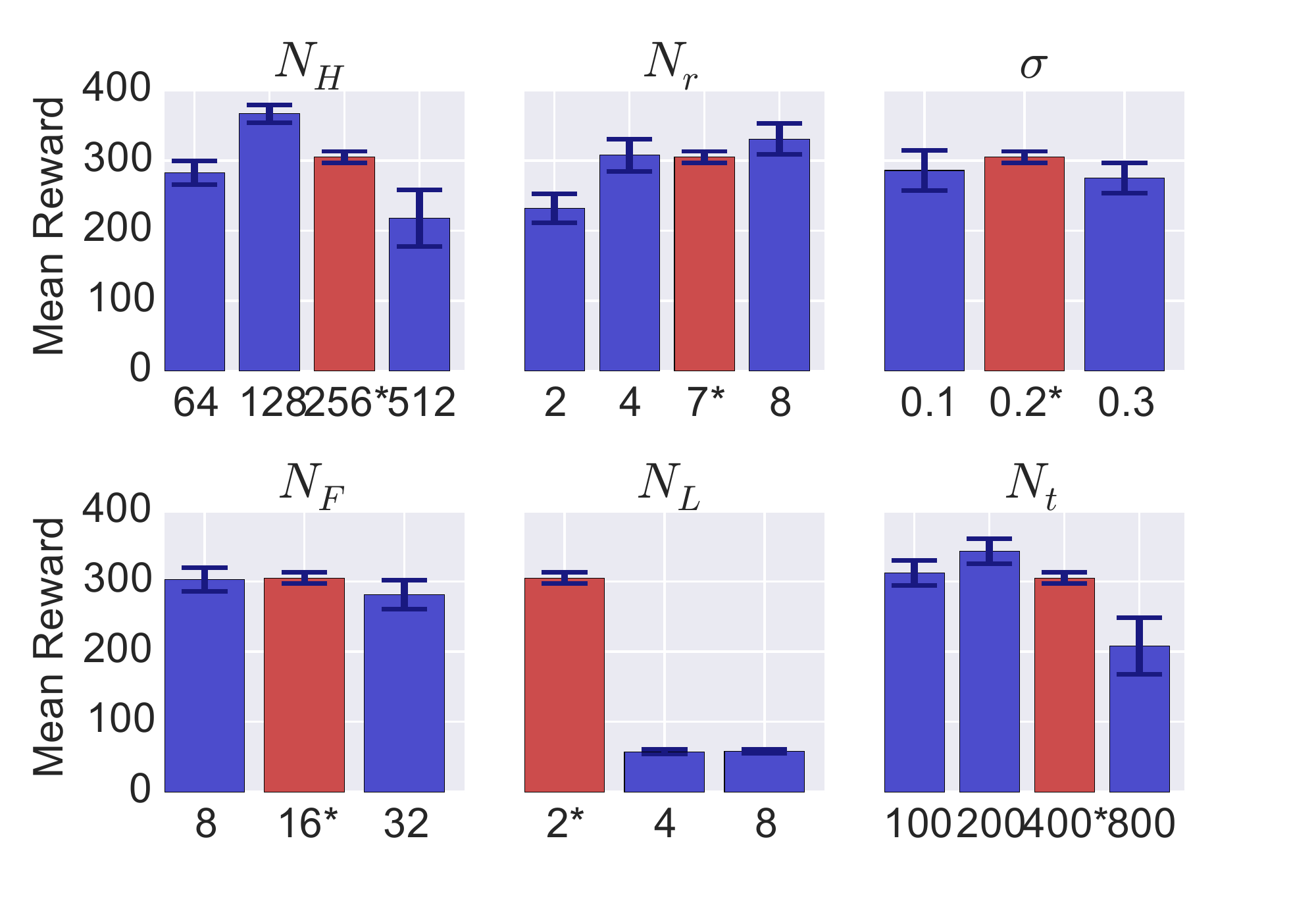}
 \caption{\label{Hyperparam}Effects of varying select hyperparameters. Results are relatively insensitive to most hyperparameters, except for choice of latent space dimension $N_L$ which appears crucial. }
\end{figure}

Other hyperparameters primarily influenced the stability of learning once a solution was obtained and the time until obtaining a working solution. To quantify the sensitivity of training to hyperparameters, we measure the average reward during the first 125 episodes of training (Fig.~\ref{Hyperparam}).

\textbf{Comparison to benchmarks}:~A survey of AI Gym's leaderboard for Cartpole-v0 and Cartpole-v1 at the time of writing suggest a range of non-zero training times between 66 epochs (\url{https://gym.openai.com/evaluations/eval_TJ422PAGRzu0XWZWiVS62A}, with the first 200 being at 15 episodes) and around 4000 episodes, although not all of these entries are documented in detail. The fastest documented algorithm at time of writing was (\url{https://gym.openai.com/evaluations/eval_MdnEPJpuRsamyDxKFZLjA}), which solves Cartpole-v1 in 67 episodes and has a first-success around 35 episodes using the method of (\url{https://webdocs.cs.ualberta.ca/~sutton/papers/barto-sutton-anderson-83.pdf}). Deeper models such as DQN \cite{mnih2013ataridqn} appear to need more episodes before they can complete the task (for example, \url{https://gym.openai.com/evaluations/eval_ODj0t6gLRli1ig9GU3uVXQ} appears to be a DQN agent which has first-success at 196 episodes but requires 867 episodes to train to stability; \url{https://jaromiru.com/2016/10/21/lets-make-a-dqn-full-dqn/} reports on tuning a DQN to solve Cartpole and seems to solve the task in around 500 episodes). 

\subsection{Task Transfer}

We examine the ability of this system to indirectly learn control policies for tasks which it has not directly trained on. To do this, we first train a controller on balancing the pole at the $x=0$ position for 125 episodes, and then impose a time-varying reward function with a target position $x(t) = A\sin(2\pi t / T)$. Additionally, because we want to check if the agent can consistently follow the target for an extended period, we increase the length of each trial to $5000$ steps. 

\begin{figure}
 \includegraphics[width=\columnwidth]{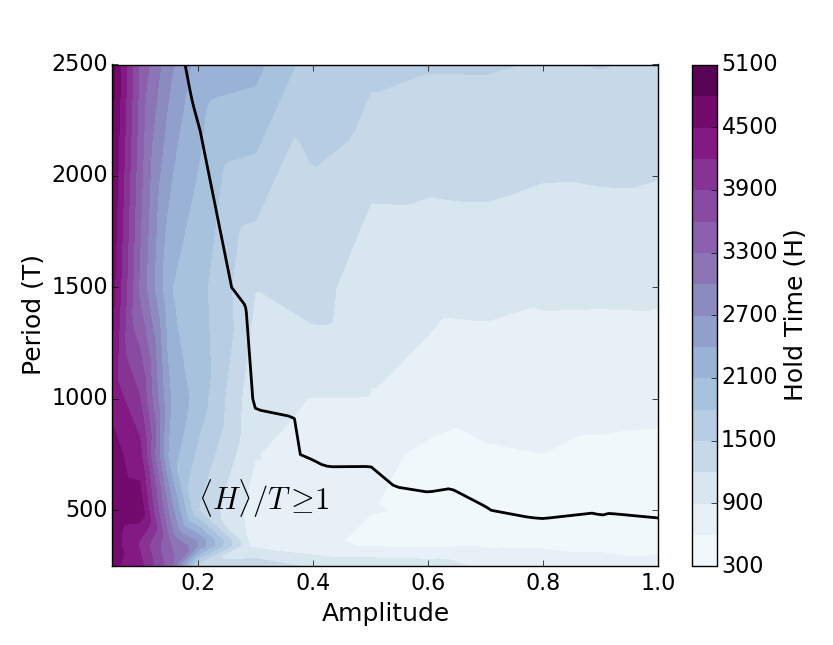}
 \caption{\label{TTHold} Average hold times over 20 trials for an agent trained to balance a stationary pole and asked to follow a varying $x$ position target. The black line indicates the region in which the agent on average could follow the target and balance the pole for more than one period of oscillation. }
\end{figure}

\begin{figure}
 \includegraphics[width=\columnwidth]{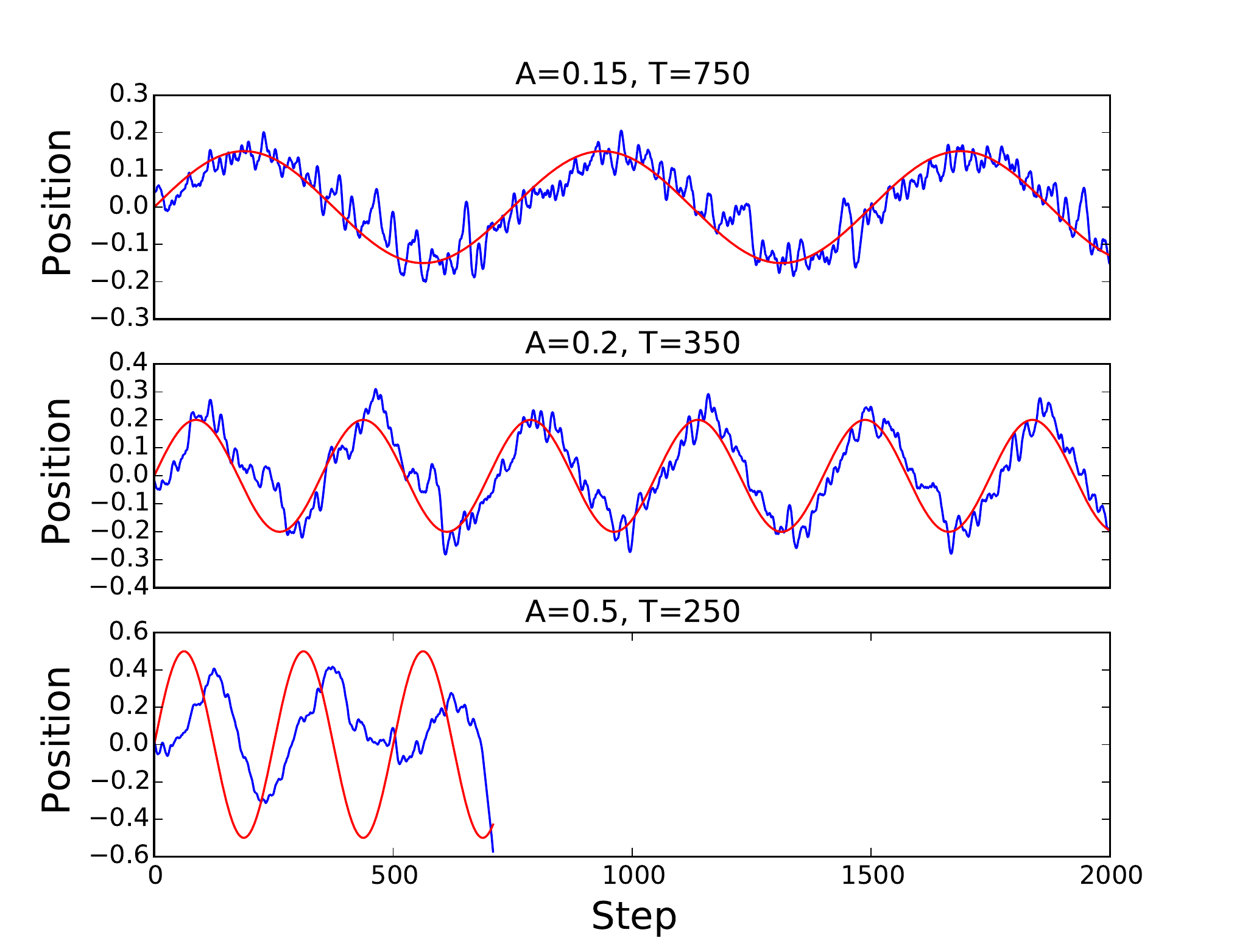}
 \caption{\label{TTTraj} Example trajectories of the cart under different reward functions than the training case. The red curve is the target position, and the blue curve is the actual trajectory followed by the cart. }
\end{figure}

We find that at worst, the agent can almost always maintain the pole in a balanced position for the same $500$ steps as in the previous task despite trying to follow the varying $x$ target (although after that it often falls). For a certain range of amplitudes and periods, the agent can successfully both follow the moving target and maintain balance for multiple periods of oscillation (Fig.~\ref{TTHold}). Example trajectories from two successful cases and one failed case are shown in (Fig.~\ref{TTTraj}). For sufficiently short periods ($T<250$) we find that the agent begins to be unable to follow the target position; as a result, the hold time increases but the actual performance on the imposed reward function degrades.

A video of the cartpole under this controller with a variable reward function is available at \url{https://youtu.be/hmkC0z8SQ08}.

\section{Conclusions}

We presented an counterfactual control algorithm that can induce control policies for free using counterfactual predictions of the joint action and state sequences in the future. Our algorithm explores the latent space of a recurrent variational autoencoder to find a suitable action sequence that is expected to lead to a desired final state. This approach allows flexible changes of goals or rewards on the fly. Our experiments show that control policies induced by a generative model are indeed capable of performing a task. Furthermore, our results indicate that counterfactual predictions into the future can be used to produce control policies to  a novel task by showing the agent can follow a continuously shifting target position despite the fact that it was never directly trained on the task.

Source code for these experiments is available at \url{https://github.com/arayabrain/GenerativeControl}.

\section*{Acknowledgements}

We would like to acknowledge Nathaniel Virgo for helpful discussions and feedback.

\bibliography{cartpole-reference}

\begin{thebibliography}{27}%
\makeatletter
\providecommand \@ifxundefined [1]{%
 \@ifx{#1\undefined}
}%
\providecommand \@ifnum [1]{%
 \ifnum #1\expandafter \@firstoftwo
 \else \expandafter \@secondoftwo
 \fi
}%
\providecommand \@ifx [1]{%
 \ifx #1\expandafter \@firstoftwo
 \else \expandafter \@secondoftwo
 \fi
}%
\providecommand \natexlab [1]{#1}%
\providecommand \enquote  [1]{``#1''}%
\providecommand \bibnamefont  [1]{#1}%
\providecommand \bibfnamefont [1]{#1}%
\providecommand \citenamefont [1]{#1}%
\providecommand \href@noop [0]{\@secondoftwo}%
\providecommand \href [0]{\begingroup \@sanitize@url \@href}%
\providecommand \@href[1]{\@@startlink{#1}\@@href}%
\providecommand \@@href[1]{\endgroup#1\@@endlink}%
\providecommand \@sanitize@url [0]{\catcode `\\12\catcode `\$12\catcode
  `\&12\catcode `\#12\catcode `\^12\catcode `\_12\catcode `\%12\relax}%
\providecommand \@@startlink[1]{}%
\providecommand \@@endlink[0]{}%
\providecommand \url  [0]{\begingroup\@sanitize@url \@url }%
\providecommand \@url [1]{\endgroup\@href {#1}{\urlprefix }}%
\providecommand \urlprefix  [0]{URL }%
\providecommand \Eprint [0]{\href }%
\providecommand \doibase [0]{http://dx.doi.org/}%
\providecommand \selectlanguage [0]{\@gobble}%
\providecommand \bibinfo  [0]{\@secondoftwo}%
\providecommand \bibfield  [0]{\@secondoftwo}%
\providecommand \translation [1]{[#1]}%
\providecommand \BibitemOpen [0]{}%
\providecommand \bibitemStop [0]{}%
\providecommand \bibitemNoStop [0]{.\EOS\space}%
\providecommand \EOS [0]{\spacefactor3000\relax}%
\providecommand \BibitemShut  [1]{\csname bibitem#1\endcsname}%
\let\auto@bib@innerbib\@empty
\bibitem [{\citenamefont {Sutton}\ and\ \citenamefont
  {Barto}(1998)}]{sutton1998reinforcement}%
  \BibitemOpen
  \bibfield  {author} {\bibinfo {author} {\bibfnamefont {R.~S.}\ \bibnamefont
  {Sutton}}\ and\ \bibinfo {author} {\bibfnamefont {A.~G.}\ \bibnamefont
  {Barto}},\ }\href@noop {} {\emph {\bibinfo {title} {Reinforcement learning:
  An introduction}}}\ (\bibinfo  {publisher} {MIT press Cambridge},\ \bibinfo
  {year} {1998})\BibitemShut {NoStop}%
\bibitem [{\citenamefont {Sutton}(1991)}]{sutton1991dyna}%
  \BibitemOpen
  \bibfield  {author} {\bibinfo {author} {\bibfnamefont {R.~S.}\ \bibnamefont
  {Sutton}},\ }\href@noop {} {\bibfield  {journal} {\bibinfo  {journal} {ACM
  SIGART Bulletin}\ }\textbf {\bibinfo {volume} {2}},\ \bibinfo {pages} {160}
  (\bibinfo {year} {1991})}\BibitemShut {NoStop}%
\bibitem [{\citenamefont {Doya}\ \emph {et~al.}(2002)\citenamefont {Doya},
  \citenamefont {Samejima}, \citenamefont {Katagiri},\ and\ \citenamefont
  {Kawato}}]{doya2002multiple}%
  \BibitemOpen
  \bibfield  {author} {\bibinfo {author} {\bibfnamefont {K.}~\bibnamefont
  {Doya}}, \bibinfo {author} {\bibfnamefont {K.}~\bibnamefont {Samejima}},
  \bibinfo {author} {\bibfnamefont {K.-i.}\ \bibnamefont {Katagiri}}, \ and\
  \bibinfo {author} {\bibfnamefont {M.}~\bibnamefont {Kawato}},\ }\href@noop {}
  {\bibfield  {journal} {\bibinfo  {journal} {Neural computation}\ }\textbf
  {\bibinfo {volume} {14}},\ \bibinfo {pages} {1347} (\bibinfo {year}
  {2002})}\BibitemShut {NoStop}%
\bibitem [{\citenamefont {Williams}(1992)}]{williams1992simple}%
  \BibitemOpen
  \bibfield  {author} {\bibinfo {author} {\bibfnamefont {R.~J.}\ \bibnamefont
  {Williams}},\ }\href@noop {} {\bibfield  {journal} {\bibinfo  {journal}
  {Machine learning}\ }\textbf {\bibinfo {volume} {8}},\ \bibinfo {pages} {229}
  (\bibinfo {year} {1992})}\BibitemShut {NoStop}%
\bibitem [{\citenamefont {Moriarty}\ \emph {et~al.}(1999)\citenamefont
  {Moriarty}, \citenamefont {Schultz},\ and\ \citenamefont
  {Grefenstette}}]{moriarty1999evolutionary}%
  \BibitemOpen
  \bibfield  {author} {\bibinfo {author} {\bibfnamefont {D.~E.}\ \bibnamefont
  {Moriarty}}, \bibinfo {author} {\bibfnamefont {A.~C.}\ \bibnamefont
  {Schultz}}, \ and\ \bibinfo {author} {\bibfnamefont {J.~J.}\ \bibnamefont
  {Grefenstette}},\ }\href@noop {} {\bibfield  {journal} {\bibinfo  {journal}
  {J. Artif. Intell. Res.(JAIR)}\ }\textbf {\bibinfo {volume} {11}},\ \bibinfo
  {pages} {241} (\bibinfo {year} {1999})}\BibitemShut {NoStop}%
\bibitem [{\citenamefont {Mnih}\ \emph {et~al.}(2015)\citenamefont {Mnih},
  \citenamefont {Kavukcuoglu}, \citenamefont {Silver}, \citenamefont {Rusu},
  \citenamefont {Veness}, \citenamefont {Bellemare}, \citenamefont {Graves},
  \citenamefont {Riedmiller}, \citenamefont {Fidjeland}, \citenamefont
  {Ostrovski} \emph {et~al.}}]{mnih2015human}%
  \BibitemOpen
  \bibfield  {author} {\bibinfo {author} {\bibfnamefont {V.}~\bibnamefont
  {Mnih}}, \bibinfo {author} {\bibfnamefont {K.}~\bibnamefont {Kavukcuoglu}},
  \bibinfo {author} {\bibfnamefont {D.}~\bibnamefont {Silver}}, \bibinfo
  {author} {\bibfnamefont {A.~A.}\ \bibnamefont {Rusu}}, \bibinfo {author}
  {\bibfnamefont {J.}~\bibnamefont {Veness}}, \bibinfo {author} {\bibfnamefont
  {M.~G.}\ \bibnamefont {Bellemare}}, \bibinfo {author} {\bibfnamefont
  {A.}~\bibnamefont {Graves}}, \bibinfo {author} {\bibfnamefont
  {M.}~\bibnamefont {Riedmiller}}, \bibinfo {author} {\bibfnamefont {A.~K.}\
  \bibnamefont {Fidjeland}}, \bibinfo {author} {\bibfnamefont {G.}~\bibnamefont
  {Ostrovski}},  \emph {et~al.},\ }\href@noop {} {\bibfield  {journal}
  {\bibinfo  {journal} {Nature}\ }\textbf {\bibinfo {volume} {518}},\ \bibinfo
  {pages} {529} (\bibinfo {year} {2015})}\BibitemShut {NoStop}%
\bibitem [{\citenamefont {Mnih}\ \emph {et~al.}(2016)\citenamefont {Mnih},
  \citenamefont {Badia}, \citenamefont {Mirza}, \citenamefont {Graves},
  \citenamefont {Lillicrap}, \citenamefont {Harley}, \citenamefont {Silver},\
  and\ \citenamefont {Kavukcuoglu}}]{mnih2016asynchronous}%
  \BibitemOpen
  \bibfield  {author} {\bibinfo {author} {\bibfnamefont {V.}~\bibnamefont
  {Mnih}}, \bibinfo {author} {\bibfnamefont {A.~P.}\ \bibnamefont {Badia}},
  \bibinfo {author} {\bibfnamefont {M.}~\bibnamefont {Mirza}}, \bibinfo
  {author} {\bibfnamefont {A.}~\bibnamefont {Graves}}, \bibinfo {author}
  {\bibfnamefont {T.~P.}\ \bibnamefont {Lillicrap}}, \bibinfo {author}
  {\bibfnamefont {T.}~\bibnamefont {Harley}}, \bibinfo {author} {\bibfnamefont
  {D.}~\bibnamefont {Silver}}, \ and\ \bibinfo {author} {\bibfnamefont
  {K.}~\bibnamefont {Kavukcuoglu}},\ }in\ \href@noop {} {\emph {\bibinfo
  {booktitle} {International Conference on Machine Learning}}}\ (\bibinfo
  {year} {2016})\BibitemShut {NoStop}%
\bibitem [{\citenamefont {Abbeel}\ and\ \citenamefont
  {Ng}(2004)}]{abbeel2004apprenticeship}%
  \BibitemOpen
  \bibfield  {author} {\bibinfo {author} {\bibfnamefont {P.}~\bibnamefont
  {Abbeel}}\ and\ \bibinfo {author} {\bibfnamefont {A.~Y.}\ \bibnamefont
  {Ng}},\ }in\ \href@noop {} {\emph {\bibinfo {booktitle} {Proceedings of the
  twenty-first international conference on Machine learning}}}\ (\bibinfo
  {organization} {ACM},\ \bibinfo {year} {2004})\ p.~\bibinfo {pages}
  {1}\BibitemShut {NoStop}%
\bibitem [{\citenamefont {Lenz}\ \emph {et~al.}(2015)\citenamefont {Lenz},
  \citenamefont {Knepper},\ and\ \citenamefont {Saxena}}]{lenz2015deepmpc}%
  \BibitemOpen
  \bibfield  {author} {\bibinfo {author} {\bibfnamefont {I.}~\bibnamefont
  {Lenz}}, \bibinfo {author} {\bibfnamefont {R.~A.}\ \bibnamefont {Knepper}}, \
  and\ \bibinfo {author} {\bibfnamefont {A.}~\bibnamefont {Saxena}},\ }in\
  \href@noop {} {\emph {\bibinfo {booktitle} {Robotics: Science and Systems}}}\
  (\bibinfo {year} {2015})\BibitemShut {NoStop}%
\bibitem [{\citenamefont {Oh}\ \emph {et~al.}(2015)\citenamefont {Oh},
  \citenamefont {Guo}, \citenamefont {Lee}, \citenamefont {Lewis},\ and\
  \citenamefont {Singh}}]{oh2015action}%
  \BibitemOpen
  \bibfield  {author} {\bibinfo {author} {\bibfnamefont {J.}~\bibnamefont
  {Oh}}, \bibinfo {author} {\bibfnamefont {X.}~\bibnamefont {Guo}}, \bibinfo
  {author} {\bibfnamefont {H.}~\bibnamefont {Lee}}, \bibinfo {author}
  {\bibfnamefont {R.~L.}\ \bibnamefont {Lewis}}, \ and\ \bibinfo {author}
  {\bibfnamefont {S.}~\bibnamefont {Singh}},\ }in\ \href@noop {} {\emph
  {\bibinfo {booktitle} {Advances in Neural Information Processing Systems}}}\
  (\bibinfo {year} {2015})\ pp.\ \bibinfo {pages} {2863--2871}\BibitemShut
  {NoStop}%
\bibitem [{\citenamefont {Fragkiadaki}\ \emph {et~al.}(2015)\citenamefont
  {Fragkiadaki}, \citenamefont {Agrawal}, \citenamefont {Levine},\ and\
  \citenamefont {Malik}}]{fragkiadaki2015learning}%
  \BibitemOpen
  \bibfield  {author} {\bibinfo {author} {\bibfnamefont {K.}~\bibnamefont
  {Fragkiadaki}}, \bibinfo {author} {\bibfnamefont {P.}~\bibnamefont
  {Agrawal}}, \bibinfo {author} {\bibfnamefont {S.}~\bibnamefont {Levine}}, \
  and\ \bibinfo {author} {\bibfnamefont {J.}~\bibnamefont {Malik}},\
  }\href@noop {} {\bibfield  {journal} {\bibinfo  {journal} {arXiv preprint
  arXiv:1511.07404}\ } (\bibinfo {year} {2015})}\BibitemShut {NoStop}%
\bibitem [{\citenamefont {Dosovitskiy}\ and\ \citenamefont
  {Koltun}(2016)}]{dosovitskiy2016learning}%
  \BibitemOpen
  \bibfield  {author} {\bibinfo {author} {\bibfnamefont {A.}~\bibnamefont
  {Dosovitskiy}}\ and\ \bibinfo {author} {\bibfnamefont {V.}~\bibnamefont
  {Koltun}},\ }\href@noop {} {\bibfield  {journal} {\bibinfo  {journal} {arXiv
  preprint arXiv:1611.01779}\ } (\bibinfo {year} {2016})}\BibitemShut {NoStop}%
\bibitem [{\citenamefont {Schmidhuber}(2010)}]{schmidhuber2010formal}%
  \BibitemOpen
  \bibfield  {author} {\bibinfo {author} {\bibfnamefont {J.}~\bibnamefont
  {Schmidhuber}},\ }\href@noop {} {\bibfield  {journal} {\bibinfo  {journal}
  {IEEE Transactions on Autonomous Mental Development}\ }\textbf {\bibinfo
  {volume} {2}},\ \bibinfo {pages} {230} (\bibinfo {year} {2010})}\BibitemShut
  {NoStop}%
\bibitem [{\citenamefont {Houthooft}\ \emph {et~al.}(2016)\citenamefont
  {Houthooft}, \citenamefont {Chen}, \citenamefont {Duan}, \citenamefont
  {Schulman}, \citenamefont {De~Turck},\ and\ \citenamefont
  {Abbeel}}]{houthooft2016vime}%
  \BibitemOpen
  \bibfield  {author} {\bibinfo {author} {\bibfnamefont {R.}~\bibnamefont
  {Houthooft}}, \bibinfo {author} {\bibfnamefont {X.}~\bibnamefont {Chen}},
  \bibinfo {author} {\bibfnamefont {Y.}~\bibnamefont {Duan}}, \bibinfo {author}
  {\bibfnamefont {J.}~\bibnamefont {Schulman}}, \bibinfo {author}
  {\bibfnamefont {F.}~\bibnamefont {De~Turck}}, \ and\ \bibinfo {author}
  {\bibfnamefont {P.}~\bibnamefont {Abbeel}},\ }in\ \href@noop {} {\emph
  {\bibinfo {booktitle} {Advances in Neural Information Processing Systems}}}\
  (\bibinfo {year} {2016})\ pp.\ \bibinfo {pages} {1109--1117}\BibitemShut
  {NoStop}%
\bibitem [{\citenamefont {Friston}\ \emph {et~al.}(2012)\citenamefont
  {Friston}, \citenamefont {Adams}, \citenamefont {Perrinet},\ and\
  \citenamefont {Breakspear}}]{friston2012perceptions}%
  \BibitemOpen
  \bibfield  {author} {\bibinfo {author} {\bibfnamefont {K.}~\bibnamefont
  {Friston}}, \bibinfo {author} {\bibfnamefont {R.}~\bibnamefont {Adams}},
  \bibinfo {author} {\bibfnamefont {L.}~\bibnamefont {Perrinet}}, \ and\
  \bibinfo {author} {\bibfnamefont {M.}~\bibnamefont {Breakspear}},\
  }\href@noop {} {\bibfield  {journal} {\bibinfo  {journal} {Frontiers in
  psychology}\ }\textbf {\bibinfo {volume} {3}},\ \bibinfo {pages} {151}
  (\bibinfo {year} {2012})}\BibitemShut {NoStop}%
\bibitem [{\citenamefont {Ho}\ and\ \citenamefont
  {Ermon}(2016)}]{ho2016generative}%
  \BibitemOpen
  \bibfield  {author} {\bibinfo {author} {\bibfnamefont {J.}~\bibnamefont
  {Ho}}\ and\ \bibinfo {author} {\bibfnamefont {S.}~\bibnamefont {Ermon}},\
  }in\ \href@noop {} {\emph {\bibinfo {booktitle} {Advances in Neural
  Information Processing Systems}}}\ (\bibinfo {year} {2016})\ pp.\ \bibinfo
  {pages} {4565--4573}\BibitemShut {NoStop}%
\bibitem [{\citenamefont {Kingma}\ and\ \citenamefont
  {Welling}(2013)}]{kingma2013vae}%
  \BibitemOpen
  \bibfield  {author} {\bibinfo {author} {\bibfnamefont {D.~P.}\ \bibnamefont
  {Kingma}}\ and\ \bibinfo {author} {\bibfnamefont {M.}~\bibnamefont
  {Welling}},\ }\href {https://arxiv.org/abs/1312.6114} {\bibfield  {journal}
  {\bibinfo  {journal} {arXiv preprint arXiv:1312.6114}\ } (\bibinfo {year}
  {2013})}\BibitemShut {NoStop}%
\bibitem [{\citenamefont {Goodfellow}\ \emph {et~al.}(2014)\citenamefont
  {Goodfellow}, \citenamefont {Pouget-Abadie}, \citenamefont {Mirza},
  \citenamefont {Xu}, \citenamefont {Warde-Farley}, \citenamefont {Ozair},
  \citenamefont {Courville},\ and\ \citenamefont {Bengio}}]{goodfellow2014gan}%
  \BibitemOpen
  \bibfield  {author} {\bibinfo {author} {\bibfnamefont {I.~J.}\ \bibnamefont
  {Goodfellow}}, \bibinfo {author} {\bibfnamefont {J.}~\bibnamefont
  {Pouget-Abadie}}, \bibinfo {author} {\bibfnamefont {M.}~\bibnamefont
  {Mirza}}, \bibinfo {author} {\bibfnamefont {B.}~\bibnamefont {Xu}}, \bibinfo
  {author} {\bibfnamefont {D.}~\bibnamefont {Warde-Farley}}, \bibinfo {author}
  {\bibfnamefont {S.}~\bibnamefont {Ozair}}, \bibinfo {author} {\bibfnamefont
  {A.}~\bibnamefont {Courville}}, \ and\ \bibinfo {author} {\bibfnamefont
  {Y.}~\bibnamefont {Bengio}},\ }\href {https://arxiv.org/abs/1406.2661}
  {\bibfield  {journal} {\bibinfo  {journal} {arXiv preprint arXiv:1406.2661}\
  } (\bibinfo {year} {2014})}\BibitemShut {NoStop}%
\bibitem [{\citenamefont {Guttenberg}\ \emph {et~al.}(2016)\citenamefont
  {Guttenberg}, \citenamefont {Sinapayen}, \citenamefont {Yu}, \citenamefont
  {Virgo},\ and\ \citenamefont {Kanai}}]{guttenberg2016recurrent}%
  \BibitemOpen
  \bibfield  {author} {\bibinfo {author} {\bibfnamefont {N.}~\bibnamefont
  {Guttenberg}}, \bibinfo {author} {\bibfnamefont {L.}~\bibnamefont
  {Sinapayen}}, \bibinfo {author} {\bibfnamefont {Y.}~\bibnamefont {Yu}},
  \bibinfo {author} {\bibfnamefont {N.}~\bibnamefont {Virgo}}, \ and\ \bibinfo
  {author} {\bibfnamefont {R.}~\bibnamefont {Kanai}},\ }\href
  {http://www.araya.org/archives/1332} {\enquote {\bibinfo {title} {Recurrent
  generative auto-encoders and novelty search},}\ } (\bibinfo {year}
  {2016})\BibitemShut {NoStop}%
\bibitem [{\citenamefont {Creswell}\ \emph {et~al.}(2016)\citenamefont
  {Creswell}, \citenamefont {Arulkumaran},\ and\ \citenamefont
  {Bharath}}]{creswell2016improving}%
  \BibitemOpen
  \bibfield  {author} {\bibinfo {author} {\bibfnamefont {A.}~\bibnamefont
  {Creswell}}, \bibinfo {author} {\bibfnamefont {K.}~\bibnamefont
  {Arulkumaran}}, \ and\ \bibinfo {author} {\bibfnamefont {A.~A.}\ \bibnamefont
  {Bharath}},\ }\href@noop {} {\bibfield  {journal} {\bibinfo  {journal} {arXiv
  preprint arXiv:1610.09296}\ } (\bibinfo {year} {2016})}\BibitemShut {NoStop}%
\bibitem [{\citenamefont {Zheng}\ \emph {et~al.}(2015)\citenamefont {Zheng},
  \citenamefont {Jayasumana}, \citenamefont {Romera-Paredes}, \citenamefont
  {Vineet}, \citenamefont {Su}, \citenamefont {Du}, \citenamefont {Huang},\
  and\ \citenamefont {Torr}}]{zheng2015conditional}%
  \BibitemOpen
  \bibfield  {author} {\bibinfo {author} {\bibfnamefont {S.}~\bibnamefont
  {Zheng}}, \bibinfo {author} {\bibfnamefont {S.}~\bibnamefont {Jayasumana}},
  \bibinfo {author} {\bibfnamefont {B.}~\bibnamefont {Romera-Paredes}},
  \bibinfo {author} {\bibfnamefont {V.}~\bibnamefont {Vineet}}, \bibinfo
  {author} {\bibfnamefont {Z.}~\bibnamefont {Su}}, \bibinfo {author}
  {\bibfnamefont {D.}~\bibnamefont {Du}}, \bibinfo {author} {\bibfnamefont
  {C.}~\bibnamefont {Huang}}, \ and\ \bibinfo {author} {\bibfnamefont {P.~H.}\
  \bibnamefont {Torr}},\ }in\ \href@noop {} {\emph {\bibinfo {booktitle}
  {Proceedings of the IEEE International Conference on Computer Vision}}}\
  (\bibinfo {year} {2015})\ pp.\ \bibinfo {pages} {1529--1537}\BibitemShut
  {NoStop}%
\bibitem [{\citenamefont {Chen}\ \emph {et~al.}(2016)\citenamefont {Chen},
  \citenamefont {Duan}, \citenamefont {Houthooft}, \citenamefont {Schulman},
  \citenamefont {Sutskever},\ and\ \citenamefont {Abbeel}}]{chen2016infogan}%
  \BibitemOpen
  \bibfield  {author} {\bibinfo {author} {\bibfnamefont {X.}~\bibnamefont
  {Chen}}, \bibinfo {author} {\bibfnamefont {Y.}~\bibnamefont {Duan}}, \bibinfo
  {author} {\bibfnamefont {R.}~\bibnamefont {Houthooft}}, \bibinfo {author}
  {\bibfnamefont {J.}~\bibnamefont {Schulman}}, \bibinfo {author}
  {\bibfnamefont {I.}~\bibnamefont {Sutskever}}, \ and\ \bibinfo {author}
  {\bibfnamefont {P.}~\bibnamefont {Abbeel}},\ }\href
  {https://arxiv.org/abs/1606.03657} {\bibfield  {journal} {\bibinfo  {journal}
  {arXiv preprint arXiv:1606.03657}\ } (\bibinfo {year} {2016})},\ \Eprint
  {http://arxiv.org/abs/1606.03657} {1606.03657} \BibitemShut {NoStop}%
\bibitem [{\citenamefont {Mathieu}\ \emph {et~al.}(2016)\citenamefont
  {Mathieu}, \citenamefont {Zhao}, \citenamefont {Sprechmann}, \citenamefont
  {Ramesh},\ and\ \citenamefont {LeCun}}]{mathieu2016dr}%
  \BibitemOpen
  \bibfield  {author} {\bibinfo {author} {\bibfnamefont {M.}~\bibnamefont
  {Mathieu}}, \bibinfo {author} {\bibfnamefont {J.}~\bibnamefont {Zhao}},
  \bibinfo {author} {\bibfnamefont {P.}~\bibnamefont {Sprechmann}}, \bibinfo
  {author} {\bibfnamefont {A.}~\bibnamefont {Ramesh}}, \ and\ \bibinfo {author}
  {\bibfnamefont {Y.}~\bibnamefont {LeCun}},\ }\href
  {https://arxiv.org/abs/1611.03383} {\bibfield  {journal} {\bibinfo  {journal}
  {arXiv preprint arXiv: 1611.03383}\ } (\bibinfo {year} {2016})},\ \Eprint
  {http://arxiv.org/abs/1611.03383} {1611.03383} \BibitemShut {NoStop}%
\bibitem [{\citenamefont {Brockman}\ \emph {et~al.}(2016)\citenamefont
  {Brockman}, \citenamefont {Cheung}, \citenamefont {Pettersson}, \citenamefont
  {Schneider}, \citenamefont {Schulman}, \citenamefont {Tang},\ and\
  \citenamefont {Zaremba}}]{brockman2016gym}%
  \BibitemOpen
  \bibfield  {author} {\bibinfo {author} {\bibfnamefont {G.}~\bibnamefont
  {Brockman}}, \bibinfo {author} {\bibfnamefont {V.}~\bibnamefont {Cheung}},
  \bibinfo {author} {\bibfnamefont {L.}~\bibnamefont {Pettersson}}, \bibinfo
  {author} {\bibfnamefont {J.}~\bibnamefont {Schneider}}, \bibinfo {author}
  {\bibfnamefont {J.}~\bibnamefont {Schulman}}, \bibinfo {author}
  {\bibfnamefont {J.}~\bibnamefont {Tang}}, \ and\ \bibinfo {author}
  {\bibfnamefont {W.}~\bibnamefont {Zaremba}},\ }\href
  {https://arxiv.org/abs/1606.01540} {\bibfield  {journal} {\bibinfo  {journal}
  {arXiv preprint arXiv:1606.01540}\ } (\bibinfo {year} {2016})}\BibitemShut
  {NoStop}%
\bibitem [{\citenamefont {Minorsky}(1922)}]{minorsky1922pid}%
  \BibitemOpen
  \bibfield  {author} {\bibinfo {author} {\bibfnamefont {N.}~\bibnamefont
  {Minorsky}},\ }\href {\doibase 10.1111/j.1559-3584.1922.tb04958.x} {\bibfield
   {journal} {\bibinfo  {journal} {Journal of the American Society for Naval
  Engineers}\ }\textbf {\bibinfo {volume} {34}},\ \bibinfo {pages} {280}
  (\bibinfo {year} {1922})}\BibitemShut {NoStop}%
\bibitem [{\citenamefont {Kingma}\ and\ \citenamefont
  {Ba}(2014)}]{kingma2014adam}%
  \BibitemOpen
  \bibfield  {author} {\bibinfo {author} {\bibfnamefont {D.~P.}\ \bibnamefont
  {Kingma}}\ and\ \bibinfo {author} {\bibfnamefont {J.}~\bibnamefont {Ba}},\
  }\href {https://arxiv.org/abs/1412.6980} {\bibfield  {journal} {\bibinfo
  {journal} {arXiv preprint arXiv:1412.6980}\ } (\bibinfo {year}
  {2014})}\BibitemShut {NoStop}%
\bibitem [{\citenamefont {Mnih}\ \emph {et~al.}(2013)\citenamefont {Mnih},
  \citenamefont {Kavukcuoglu}, \citenamefont {Silver}, \citenamefont {Graves},
  \citenamefont {Antonoglou}, \citenamefont {Wierstra},\ and\ \citenamefont
  {Riedmiller}}]{mnih2013ataridqn}%
  \BibitemOpen
  \bibfield  {author} {\bibinfo {author} {\bibfnamefont {V.}~\bibnamefont
  {Mnih}}, \bibinfo {author} {\bibfnamefont {K.}~\bibnamefont {Kavukcuoglu}},
  \bibinfo {author} {\bibfnamefont {D.}~\bibnamefont {Silver}}, \bibinfo
  {author} {\bibfnamefont {A.}~\bibnamefont {Graves}}, \bibinfo {author}
  {\bibfnamefont {I.}~\bibnamefont {Antonoglou}}, \bibinfo {author}
  {\bibfnamefont {D.}~\bibnamefont {Wierstra}}, \ and\ \bibinfo {author}
  {\bibfnamefont {M.}~\bibnamefont {Riedmiller}},\ }\href
  {https://arxiv.org/abs/1312.5602} {\bibfield  {journal} {\bibinfo  {journal}
  {arXiv preprint arXiv:1312.5602}\ } (\bibinfo {year} {2013})}\BibitemShut
  {NoStop}%
\end{thebibliography}%

\end{document}